\documentclass[runningheads]{llncs}

\usepackage{graphicx}

\usepackage{booktabs} 
\usepackage{enumitem} 
\usepackage{subcaption}
\usepackage{hyperref}
\usepackage{listings} 
\usepackage{orcidlink}
\usepackage{tablefootnote}
\usepackage{todonotes}
\usepackage{amsmath}
\usepackage{xspace}
\usepackage[nocompress]{cite} 

\usepackage{graphicx}
\usepackage{tikz}
\usetikzlibrary{patterns}
\usepackage{pdfpages}

\DeclareRobustCommand{\legendsquare}[1]{%
  \tikz[baseline=(a.south)]{\node[#1, inner sep=.8ex, outer sep=0] (a) {};}%
}

\newcommand{\PIF}{\textit{PIF}\xspace}
\newcommand{\UCSF}{\textit{UCSF}\xspace}

\begin{document}
\title{DocILE Benchmark for Document Information Localization and Extraction}

\ifx\review\undefined
    \author{Štěpán Šimsa\inst{1}\orcidlink{0000-0001-6687-1210} \and
    Milan Šulc  \inst{1}\orcidlink{0000-0002-6321-0131} \and
    Michal Uřičář  \inst{1}\orcidlink{0000-0002-2606-4470} \and
    Yash Patel \inst{2}\orcidlink{0000-0001-9373-529X} \and\\
    Ahmed Hamdi \inst{3}\orcidlink{0000-0002-8964-2135} \and
    Matěj Kocián  \inst{1}\orcidlink{0000-0002-0124-9348} \and
    Matyáš Skalický \inst{1}\orcidlink{0000-0002-0197-7134}
    \and
    Jiří Matas \inst{2}\orcidlink{0000-0003-0863-4844}
    \and
    Antoine Doucet \inst{3}\orcidlink{0000-0001-6160-3356}
    \and
    Mickaël Coustaty \inst{3}\orcidlink{0000-0002-0123-439X}
    \and
    Dimosthenis Karatzas \inst{4}\orcidlink{0000-0001-8762-4454}
    }
    \authorrunning{Š. Šimsa et al.}
    \institute{
    Rossum.ai,
      \url{https://rossum.ai},
      \email{\{name.surname\}@rossum.ai} \and
    Visual Recognition Group, Czech Technical University in Prague \and
    University of La Rochelle, France
    \and
    Computer Vision Center, Universitat Autónoma de Barcelona, Spain
    }
\fi
\maketitle

\setcounter{footnote}{0}

\begin{abstract}
This paper introduces the DocILE benchmark with the largest dataset of business documents for the tasks of \textit{Key Information Localization and Extraction} and \textit{Line Item Recognition}. It contains $6.7$k annotated business documents, $100$k synthetically generated documents, and nearly~$1$M unlabeled documents for unsupervised pre-training. The dataset has been built with knowledge of domain- and task-specific aspects, resulting in the following key features: (i) annotations in $55$ classes, which surpasses the granularity of previously published key information extraction datasets by a large margin; (ii) Line Item Recognition represents a highly practical information extraction task, where key information has to be assigned to items in a table; 
 (iii) documents come from numerous layouts and the test set includes zero- and few-shot cases as well as layouts commonly seen in the training set.
 The benchmark comes with several baselines, including RoBERTa, LayoutLMv3 and DETR-based Table Transformer; applied to both tasks of the DocILE benchmark, with results shared in this paper, offering a quick starting point for future work. 
\ifx\review\undefined
 The dataset, baselines and supplementary material are available at \url{https://github.com/rossumai/docile}.
\fi


\end{abstract}

\section{Introduction}

Automating information extraction from business documents has the potential to streamline repetitive human labour and allow data entry workers to focus on more strategic tasks.
Despite the recent shift towards business digitalization, the majority of Business-to-Business (B2B) communication still happens through the interchange of semi-structured\footnote{We use the term 
\textit{semi-structured documents} as~\cite{skalicky2022business,riba2019table}; visual structure is strongly related to the document semantics, but the layout is variable.} business documents such as invoices, tax forms, orders, etc. The layouts of these documents were designed for human readability, yet the downstream applications (i.e. accounting software) depend on data in a structured, computer-readable format. Traditionally, this has been solved by manual data entry, requiring substantial time to process each document. The automated process of data extraction from such documents goes far beyond Optical Character Recognition (OCR) as it requires understanding of semantics, layout and context of the information within the document.  The machine learning field dealing with this is called \emph{Document Information Extraction} (IE), a sub-category of \emph{Document Understanding} (DU).

Information Extraction from business documents lacks practical large-scale benchmarks,
as noted in~\cite{palm2017cloudscan,sunder2019one,dhakal2019one,krieger2021information,skalicky2022business}. While there are several public datasets for document understanding, as reviewed in Section~\ref{sec:related_datasets}, only a few of them focus on information extraction from business documents. 
They are typically small-scale~\cite{medvet2011probabilistic,wang2021vies,sun2021spatial}, focusing solely on receipts~\cite{park2019cord,sun2021spatial}, or limit the task, e.g., to \emph{Named Entity Recognition} (NER), 
missing location annotation 
\cite{stanislawek2021kleister,huang2019icdar2019,borchmann2021due,deepform2020}. 
Many results in the field are therefore published on private datasets~\cite{katti2018chargrid,denk2019bertgrid,holt2018extracting,palm2017cloudscan,hamdi2021information}, limiting the reproducibility and hindering further research. Digital semi-structured documents often contain sensitive information, such as names and addresses, which hampers the creation of sufficiently-large public datasets and benchmarks.

The standard problem of \emph{Key Information Extraction} (KIE) should be distinguished~\cite{skalicky2022business} from \emph{Key Information Localization and Extraction} (KILE) as the former lacks the positional information, required for effective human-in-the loop verification of the extracted data. 

Business documents often come with a list of items, e.g. a table of invoiced goods and services, where each item is represented by a set of key information, such as name, quantity and price. Extraction of such items is the target of the \emph{Line Item Recognition} (LIR) \cite{skalicky2022business}, which was not explicitly targeted by existing benchmarks.

In this work, we present the DocILE
(\textbf{Doc}ument \textbf{I}nformation \textbf{L}ocalization and \textbf{E}xtraction) 
dataset and benchmark with the following contributions:\\
(i) the largest dataset for KILE and LIR from semi-structured business documents both in terms of the number of labeled documents and categories; 
(ii) rich set of document layouts, including layout cluster annotations for all labeled documents; 
(iii) the synthetic subset being the first large synthetic dataset with KILE and LIR labels; 
(iv) detailed information about the document selection, processing and  annotations, which took around $2,500$ hours of annotation time; 
(v) baseline evaluations of popular architectures for language modelling, visually-rich document understanding and computer vision; 
(vi) is used both for a research competition, as well as a long-term benchmark of key information extraction and localization, and line item recognition systems; 
(vii) can serve other areas of research thanks to the rich annotations (table structure, layout clusters, metadata, and the HTML sources for synthetic documents).

The paper is structured as follows: Section~\ref{sec:related_work} reviews the related work. The DocILE dataset is introduced and its characteristics and its collection are described in Section~\ref{sec:docileDataset}. Section~\ref{sec:benchmark} follows with the tasks and evaluation metrics. Baseline methods are described and experimented in Section~\ref{sec:baselines}. Finally, conclusions are drawn in Section~\ref{sec:conclusions}.

\section{Related Work}
\label{sec:related_work}

To address the related work, we first introduce general approaches to document understanding, before specifically focusing on information extraction tasks and existing datasets.

\subsection{Methods for Document Understanding}
Approaches to document understanding have used various combinations of input modalities (text, spatial layout, image) to extract information from structurally rich documents. Such approaches have been successfully applied to understanding of forms~\cite{davis2019deep,hammami2015one,zhou2016irmp}, receipts~\cite{hong2022bros,huang2019icdar2019}, tables~\cite{herzig2020tapas,schreiber2017deepdesrt,zhong2019publaynet}, or invoices~\cite{lohani2019invoice,majumder2020representation,riba2019table}. 

Convolutional neural networks based approaches such as~\cite{lin2021vibertgrid,katti2018chargrid} use character or word vector-based representations to make a grid-style prediction similar to semantic segmentation. The pixels are classified into the field types for invoice documents. LayoutLM~\cite{xu2020layoutlm} modifies the BERT~\cite{devlin2018bert} language model to incorporate document layout information and visual features. The layout information is passed in the form of $2$D spatial coordinate embeddings, and the visual features for each word token are obtained via Faster-RCNN~\cite{ren2015faster}. LayoutLMv2~\cite{xu2020layoutlmv2} treats visual tokens separately, instead of adding them to the text tokens, and incorporates additional pre-training tasks. LayoutLMv3~\cite{huang2022layoutlmv3} introduces more pre-training tasks such as masked image modeling, or word-patch alignment. BROS~\cite{hong2022bros} also uses a BERT-based text encoder equipped with SPADE~\cite{hwang2020spatial} based graph classifier to predict the entity relations between the text tokens. Document understanding has also been approached from a question-answering perspective~\cite{mathew2021docvqa,mathew2022infographicvqa}. Layout-T5~\cite{tanaka2021visualmrc} uses the layout information with the generative T5~\cite{raffel2020exploring} language model, and TILT~\cite{powalski2021going} uses convolutional features with the T5 model. In UDOP~\cite{tang2022unifying}, several document understanding tasks are formulated as sequence-to-sequence modelling in a unified framework. Recently, GraphDoc~\cite{zhang2022multimodal}, a model based on graph attention networks pre-trained only on $320$k documents, has been introduced for document understanding tasks, showing satisfactory results.

Transformer-based approaches typically rely on large-scale pre-training on unlabeled documents while the fine-tuning of a specific downstream task is sufficient with much smaller annotated datasets. Noticeable amount of papers have focused on the pre-training aspect of document understanding~\cite{appalaraju2021docformer,garncarek2021lambert,gu2021unidoc,gu2022xylayoutlm,hong2022bros,lee2022formnet,li2021structext,li2021structurallm,powalski2021going}. In this paper we use the popular methods~\cite{huang2022layoutlmv3,carion2020detr,liu2019roberta} to provide the baselines for KILE and LIR on the proposed DocILE dataset.

\subsection{Information Extraction Tasks and Datasets}
\label{sec:related_datasets}
Extraction of information from documents includes many tasks and problems from basic OCR~\cite{du2020ppocr,smith2007overview,hamad2016detailed,islam2017survey,memon2020handwritten,olejniczak2023text} up to visual question answering (VQA)~\cite{mathew2021docvqa,mathew2022infographicvqa}. The landscape of IE problems and datasets was recently reviewed by Borchmann et al.~\cite{borchmann2021due}, building the DUE Benchmark for a wide range of document understanding tasks, and by Skalický et al. ~\cite{skalicky2022business}, who argue that the crucial problems for automating B2B document communication are Key Information Localization and Extraction and Line Item Recognition.

\textbf{Key Information Extraction} (KIE)~\cite{garncarek2021lambert,stanislawek2021kleister,huang2019icdar2019} aims to extract pre-defined key information (categories of "fields" -- name, email, the amount due, etc.) from a document. A number of datasets for KIE are publicly available~\cite{huang2019icdar2019,sun2021spatial,sun2021spatial,medvet2011probabilistic,wang2021vies,deepform2020,stanislawek2021kleister}. However, as noted by~\cite{skalicky2022business}, most of them are relatively small and contain only a few annotated field categories.

\textbf{Key Information Localization and Extraction} (KILE)~\cite{skalicky2022business} additionally requires precise localization of the extracted information in the input image or PDF, which is crucial for human-in-the-loop interactions, auditing, and other processing of the documents. However, many of the existing KIE datasets miss the localization annotations~\cite{stanislawek2021kleister,huang2019icdar2019,borchmann2021due}. Publicly available KILE datasets on business documents~\cite{park2019cord,sun2021spatial,wang2021vies,medvet2011probabilistic} and their sizes are listed in Table~\ref{table:datasets}. 
Due to the lack of large-scale datasets for KILE from business documents, noted by several authors~\cite{palm2017cloudscan,sunder2019one,dhakal2019one,krieger2021information,skalicky2022business}, many research publications use private datasets~\cite{katti2018chargrid,denk2019bertgrid,holt2018extracting,palm2019attend,palm2017cloudscan,liu2016unstructured,DBLP:journals/corr/abs-1903-12363,schuster2013intellix}.

\begin{table}[tb]
\setlength\tabcolsep{.8mm}
\centering
\caption{Datasets with KILE and LIR annotations for semi-structured business documents.} 

\resizebox{\textwidth}{!}{
\begin{tabular}{p{23.5mm}p{16mm}
p{11mm}p{11mm}p{11mm}p{8mm}cp{10.5mm}} 
\hline
name                                              & document type      & \# docs\newline labeled
& classes  & source    & multi page & lang. & task  \\ 
\hline
DocILE \textit{(ours)} & invoice-like & $\mathbf{106 680}$ & $\mathbf{55}$
& digital, scan & yes & en & KILE, LIR
\\ \hline
CORD              \cite{park2019cord}             & receipts           & $11 000$ & $30-42$\tablefootnote{54 classes mentioned in \cite{park2019cord}, but the repository \url{https://github.com/clovaai/cord} only considers 30 out of 42 listed classes, as of January 2023.}\newline
      & photo  & no        & id
 & $\approx$KILE, $\approx$LIR\tablefootnote{COORD annotations contain classification of word tokens (as in NER) but with the additional information which tokens are grouped together into fields or menu items, effectively upgrading the annotations to KILE/LIR field annotations.}
      \\ \hline
WildReceipt       \cite{sun2021spatial}           & receipts           & $1 740$  & $25$        & photo  & no        & en    & KILE \\ \hline
EPHOIE            \cite{wang2021vies}             & chinese forms      & $1 494$  & $10$        & scan   & no        & zh    & KILE \\ \hline
Ghega             \cite{medvet2011probabilistic}  & patents, datasheets & $246$    & $11$/$8$
& scan   & yes       & en    & KILE
\\
\hline
\end{tabular}
}
\label{table:datasets}
\end{table}

\begin{figure}[tb]
\centering
  \includegraphics[width=0.45\textwidth]{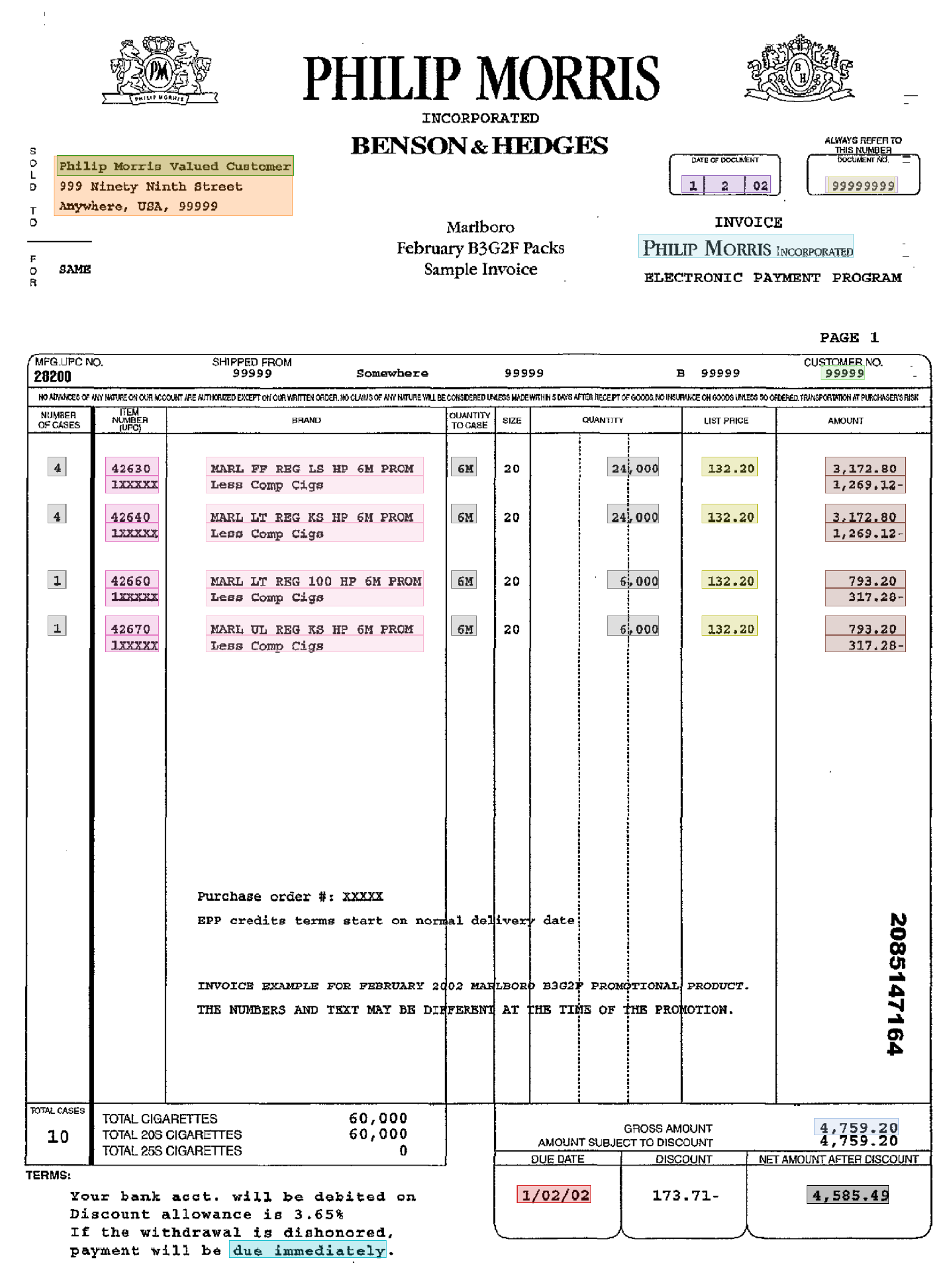}
  \hglue 0.05\textwidth 
  \includegraphics[width=0.45\textwidth]{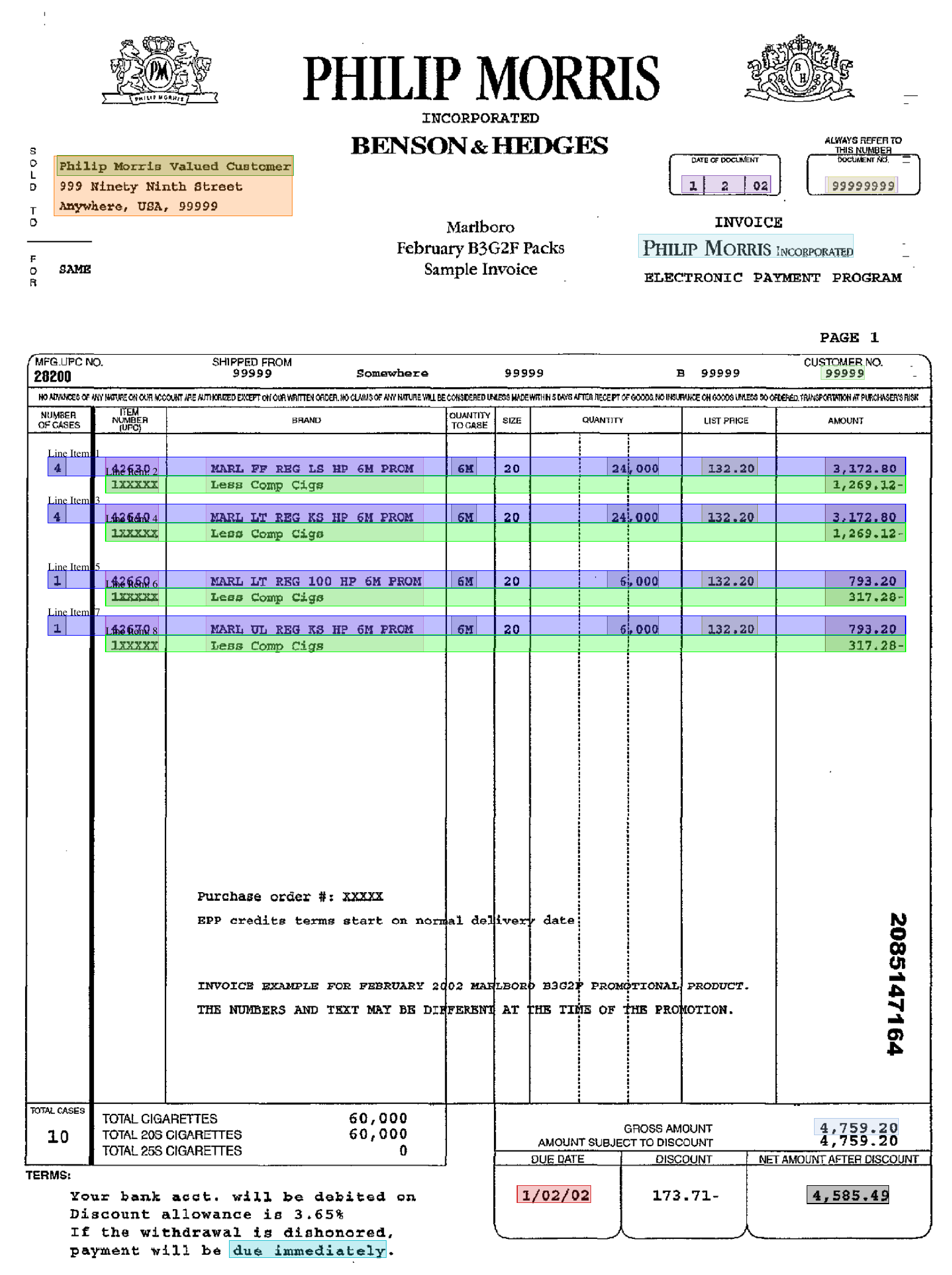}

  \includegraphics[width=\textwidth]{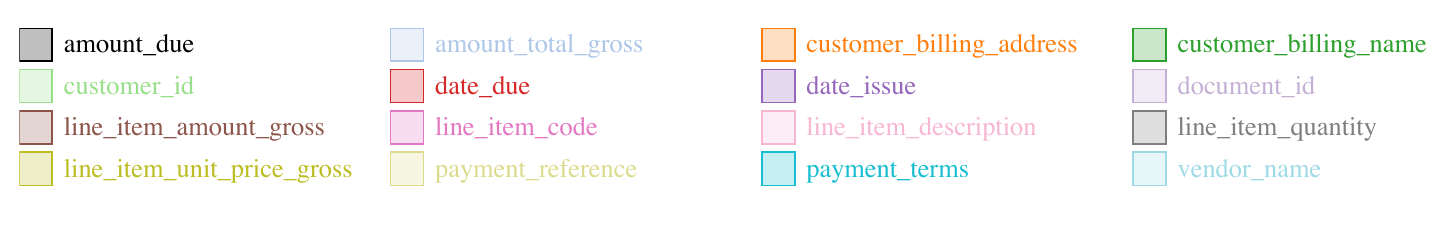}
  \caption{DocILE: a document with KILE and LIR annotations (left) and the Line Item areas emphasized (right) by alternating blue \legendsquare{draw=blue,fill=blue!40,thick}  and green \legendsquare{draw=green,fill=green!30,thick} for odd and even items, respectively. Bottom: color legend for the KILE and LIR classes.}
  \label{fig:example_document}
  \vspace{-1.5ex}
\end{figure}

\textbf{Line Item Recognition} (LIR)~\cite{skalicky2022business} is a part  of table extraction~\cite{denk2019bertgrid,holevcek2019table,palm2017cloudscan,majumder2020representation,BenschPS21} that aims at finding Line Items (LI), localizing and extracting key information for each item. The task is related to Table Structure Recognition~\cite{schreiber2017deepdesrt,tensmeyer2019deep,smock2022pubtables}, which typically aims at detecting table rows, columns and cells. 
However, sole table structure recognition is not sufficient for LIR: an enumerated item may span several rows in a table; and columns are often not sufficient to distinguish all semantic information. There are several datasets~\cite{FangTTQL12,SiegelLPA18,zhong2019publaynet,smock2022pubtables,ZhengB0ZW21,abs-2203-01017} for Table Detection and/or Structure Recognition, PubTables-1M~\cite{smock2022pubtables} being the largest with a million tables from scientific articles. 
The domain of scientific articles is prevailing among the datasets~\cite{FangTTQL12,SiegelLPA18,zhong2019publaynet,smock2022pubtables}, due to easily obtainable annotations 
from the \LaTeX\xspace source codes. However, there is a non-trivial domain shift introduced by the difference in the Tables from scientific papers and business documents.
FinTabNet~\cite{ZhengB0ZW21} and SynthTabNet~\cite{abs-2203-01017} are closer to our domain, covering table structure recognition of complex financial tables.
These datasets, however, only contain annotations of the table grid/cells. From the available datasets, CORD~\cite{park2019cord} is the closest to the task of Line Item Recognition with its annotation of sub-menu items. The documents in CORD are all receipts, which generally have simpler structure than other typical business documents, which makes the task too simple as previously mentioned in \cite{borchmann2021due}.

\textbf{Named Entity Recognition} (NER) \cite{li2020survey} is the task
of assigning one of the pre-defined categories to entities (usually words or word-pieces in the document) which makes it strongly related to KILE and LIR, especially when these entities have a known location. 
Note that the task of NER is less general as it only operates on word/token level, and using it to solve KILE is not straightforward, as the classified tokens have to be correctly aggregated into fields and fields do not necessarily have to contain whole word-tokens.

\section{The DocILE Dataset}
\label{sec:docileDataset}

In this section, we describe the DocILE dataset content and creation.

\subsection{The Annotated, the Unlabeled, and the Synthetic}
The DocILE dataset and benchmark is composed of three subsets:
\begin{enumerate}[noitemsep, topsep=3pt]
    \item an \emph{annotated set} of $6,680$ real business documents from publicly available sources which were annotated as described in Section \ref{sec:annotation}.
    \item an \emph{unlabeled set} of $932$k real business documents from publicly available sources, which can be used for unsupervised (pre-)training.
    \item a \emph{synthetic set} of $100$k documents with full task labels generated with a proprietary document generator using layouts inspired by $100$ fully annotated real business documents from the \emph{annotated set}. 
    
\end{enumerate}

The \emph{labeled} (i.e., \emph{annotated} and \emph{synthetic}) subsets contain annotations for the tasks of \emph{Key Information Localization and Extraction} and \emph{Line Item Recognition}, described below in Sections~\ref{sec:track1_kile} and~\ref{sec:track2_lir}, respectively. An example document with such annotations is shown in Figure~\ref{fig:example_document}. Table~\ref{tab:dataset-size} shows the size of the dataset. 

\begin{table}[b]
    \setlength\tabcolsep{3mm}
    \centering
    \caption{DocILE dataset --- the three subsets.}
    \label{tab:dataset-size}
    \begin{tabular}{@{}lrrr@{}}
    \toprule
                            & \textbf{annotated} & \textbf{synthetic}    & \textbf{unlabeled}    \\ \midrule
    \textbf{documents}      & 6 680          & 100 000               & 932 467               \\
    \textbf{pages}          & 8 715          & 100 000               & 3.4M                  \\
    \textbf{layout clusters}& 1 152          & 100                   & \textit{Unknown}      \\
    \textbf{pages per doc.} & 1–3           & 1                     & 1–884                 \\ \bottomrule
    \end{tabular}
    \vspace{-6mm}
\end{table}

\begin{figure}[b]
    \hspace{-5mm}
    \includegraphics[width=1.05\textwidth]{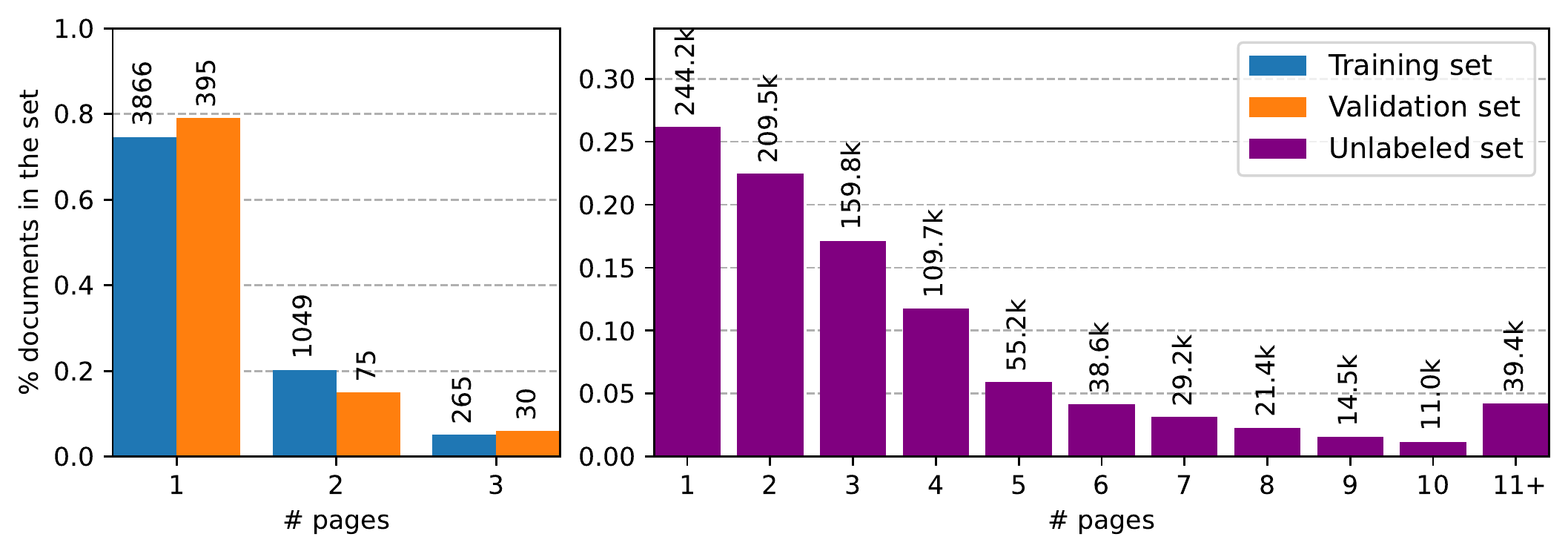}

    \vspace{-3mm}
  	\caption{Distribution of the number of document pages in the training, validation and unlabeled sets. The numbers of documents are displayed above the bars.}
  	\label{fig:npages}
\end{figure}

\begin{figure}[tb]
    \hspace{-5mm}
    \includegraphics[width=1.05\textwidth]{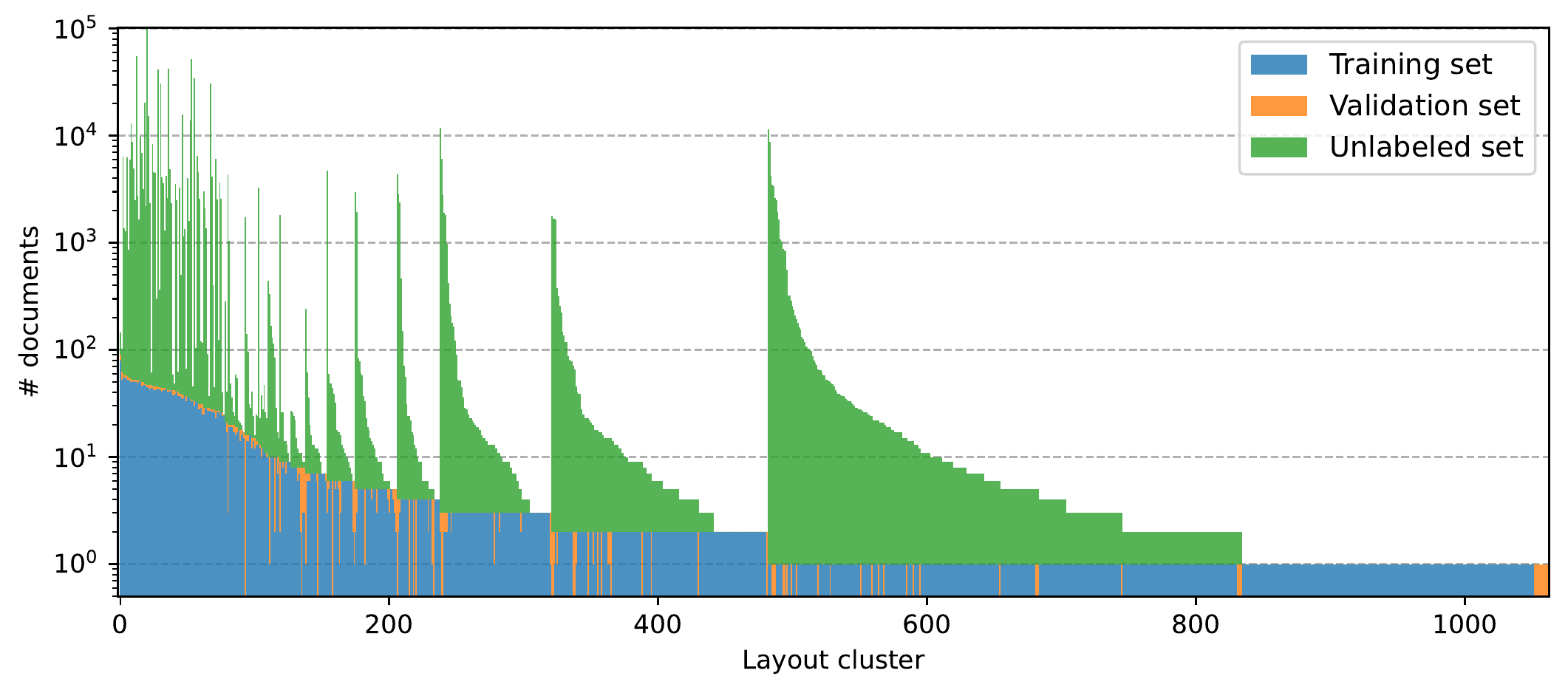}

    \vspace{-3mm}
  	\caption{The number of documents of each layout cluster in the training, validation and unlabeled sets, on a logarithmic scale. While some clusters have up to 100k documents, the largest cluster in \textit{train. + val.} contains only~90 documents.} 
  	\label{fig:docs_per_layout}
\end{figure}

\subsection{Data Sources}
\label{sec:data_sources}
Documents in the DocILE dataset come from two public data sources: UCSF Industry Documents Library~\cite{IndustryDocumentsLibrary} and Public Inspection Files (PIF)~\cite{PublicInspectionFiles}. The UCSF Industry Documents Library contains documents from industries that influence public health, such as tobacco companies.  This source has been used to create the following document datasets: RVL-CDIP~\cite{harley2015icdar}, IIT-CDIP~\cite{lewis2006building}, FUNSD~\cite{jaume2019}, DocVQA~\cite{mathew2021docvqa} and OCR-IDL~\cite{biten2022ocr}. PIF contains a variety of information about American broadcast stations. We specifically use the "political files" with documents (invoices, orders, "contracts") from TV and radio stations for political campaign ads, previously used to create the Deepform~\cite{deepform2020}. Documents from both sources were retrieved in the PDF format.

Documents for DocILE were selected from the two sources as follows.
For UCSF IDL, we used the public API~\cite{IndustryDocumentsLibraryAPI} to retrieve only publicly available documents of type \texttt{invoice}. For documents from PIF, we retrieved all "political files" from \texttt{tv}, \texttt{fm} and \texttt{am} broadcasts. We discarded documents with broken PDFs, duplicates\footnote{Using hash of page images to capture duplicates differing only in PDF metadata.}, and documents not classified as \textit{invoice-like}\footnote{\label{footnote:invoice-like}Invoice-like documents are tax invoice, order, purchase order, receipt, sales order, proforma invoice, credit note, utility bill and debit note. 
\ifx\review\undefined
We used a proprietary document-type classifier provided by \href{http://rossum.ai}{Rossum.ai}. 
\else
We used a proprietary document-type classifier.
\fi
}.
Other types of documents, such as budgets or financial reports, were discarded as they typically contain different key information.
We refer to the selected documents from the two sources as \PIF and \UCSF documents.

\subsection{Document Selection and Annotation}
\label{sec:annotation}
To capture a rich distribution of documents and make the dataset easy to work with, expensive manual annotations were only done for documents which are:
\begin{enumerate}
    \item short (1-3 pages), to annotate many different documents rather than a few long ones;
    \item written in English, for consistency and because the language distribution in the selected data sources is insufficient to consider multilingual analysis;
    \item dated\footnote{The document date was retrieved from the UCSF IDL metadata. Note that the majority of the documents in this source are from the 20th century.} 1999 or later in \UCSF, as older documents differ from the more recent ones (typewritten, etc.);
    \item representing a rich distribution of layout clusters, as shown in Figure~\ref{fig:docs_per_layout}.
\end{enumerate}

We clustered the document layouts\footnote{We loosely define layout as the positioning of fields of each type in a document. We allow, e.g., different length of values, missing values, and resulting translations of whole sections.}
based on the location of fields detected by a proprietary model for KILE. 

The clustering was manually corrected for the annotated set. 
More details about the clustering can be found in the Supplementary Material.

In the annotation process, documents were skipped if they were not \emph{invoice-like}, if they contained handwritten or redacted key information or if they listed more than one set of line items (e.g. several tables listing unrelated types of items). 

Additionally, PDF files composed of several documents (e.g., a different invoice on each page) were split and annotated separately.

For KILE and LIR, \emph{fields} are annotated as a triplet of \emph{location} (bounding box and page), \emph{field type} (class) and \emph{text}. 
LIR fields additionally contain the \emph{line item ID}, assigning the line item they belong to. If the same content is listed in several tables with different granularity (but summing to the same total amount), the less detailed set of line items is annotated.

Notice that the fields can overlap, sometimes completely. 
A field can be multi-line or contain only parts of words. 
There can be multiple fields with the same field type on the same page, either having the same value in multiple locations or even having different values as well as multiple fields with the same field type in the same line item. 
The full list of field types and their description are in the Supplementary Material.

Additional annotations, not necessary for the benchmark evaluation, are available and can be used in the training or for other research purposes. 
Table structure annotations include: 1) line item headers, representing the headers of columns corresponding to one field type in the table, and 2) the  table grid, containing information about rows, their position and classification into header, data, gap, etc., and columns, their position and field type when the values in the column correspond to this field type. Additionally, metadata contain: document type, currency, layout cluster ID, source and original filename (linking the document to the source), page count and page image sizes.

Annotating the  $6,680$ documents took approx. $2,500$ hours of annotators' time including the verification. Of the annotated documents, $53.7\%$ originate from \PIF and the remaining $46.3\%$ from \UCSF IDL. 
The annotated documents underwent the image pre-processing described in the Supplementary Material. 
All remaining documents from \PIF and \UCSF form the \emph{unlabeled} set.

\subsection{Dataset Splits}
\label{sec:dataset_splits}
The \textit{annotated} documents in the DocILE dataset are split into \emph{training} ($5,180$), \emph{validation} ($500$), and \emph{test} ($1,000$) sets.
The \emph{synthetic} set with 100k documents and \emph{unlabeled} set with 932k documents are provided as an optional extension to the training set, as  unsupervised pre-training~\cite{abs-2104-08836} and synthetic training data~\cite{gupta2016synthetic,abs-2203-01017,dosovitskiy2015flownet,nayef2019icdar2019,buvsta2019e2e} have been demonstrated to improve results of machine learning models in different domains.

The training, validation and test splitting was done so that the validation and test sets contain $25\%$ of zero-shot samples (from layouts unseen during training\footnote{For the test set, documents in both training and validation sets are considered as seen during training. Note that some test set layouts may be present in the validation set, not the training set.}), $25\%$ of few-shot samples (from layouts with $\leq 3$ documents seen during training) and $50\%$ of many-shot samples (from layouts with more examples seen during training). This allows to measure both the generalization of the evaluated methods and the advantage of observing documents of known layouts.

The test set annotations are not public and the test set predictions will be evaluated through the RRC website\footnote{\url{https://rrc.cvc.uab.es/}}, where the benchmark and competition is hosted. The validation set can be used when access to annotations and metadata is needed for experiments in different tasks.

As inputs to the document synthesis described in Section~\ref{sec:synthesis},  $100$ one-page documents were chosen from the training set, each from a different layout cluster. In the test, resp. validation sets, roughly half of the few-shot samples are from layouts for which synthetic documents were generated. There are no synthetic documents generated for zero-shot samples. For many-shot samples, $35-40\%$ of documents are from layouts with synthetic documents.

\subsection{Synthetic Documents Generation} 
\label{sec:synthesis}
To generate synthetic documents with realistic appearance and content, we used the following procedure:
First, a set of template documents from different layout clusters was selected, as described in Section~\ref{sec:dataset_splits}. All elements in the selected documents, including all present keys and values, notes, sections, borders, etc., were annotated with layout (bounding box), semantic (category) and text (where applicable) annotations. Such full annotations were the input to a rule-based document synthesizer, which uses a rich set of content generators\footnote{Such as generators of names, emails, addresses, bank account numbers, etc. Some utilize the Mimesis library \cite{mimesis}. Some content, such as keys, is copied from the annotated document.} to fill semantically relevant information in the annotated areas. Additionally, a style generator controls and enriches the look of the resulting documents (via font family and size, border styles, shifts of the document contents, etc.). The documents are first generated as HTML files and then rendered to PDF. The HTML source code of all generated documents is shared with the dataset and can be used for future work, e.g., for generative methods for conversion of document images into a markup language.

\subsection{Format}
The dataset is shared in the form of pre-processed\footnote{Pre-processing consists of correcting page orientation, de-skewing scanned documents and normalizing them to $150$ DPI.} document PDFs with task annotations in JSON. 
Additionally, each document comes with DocTR~\cite{doctr2021} OCR predictions with word-level text and 
location\footnote{Axis-aligned bounding boxes, optionally with additional snapping to reduce white space around word predictions, described in the Supplementary Material.}.

A python library 
{\it docile}\footnote{\url{https://github.com/rossumai/docile}} 
 is provided to ease the work with the dataset.

\section{Benchmark Tasks and Evaluation Metrics}
\label{sec:benchmark}

Sections \ref{sec:track1_kile} and \ref{sec:track2_lir} describe the two benchmark tasks as introduced in the teaser 
\cite{simsa2023docile} 
along with the challenge evaluation metrics used for the leaderboard ranking. 
Additional evaluation metrics are described in the Supplementary Material.

\subsection{Track 1: Key Information Localization and Extraction}
\label{sec:track1_kile}
The goal of the first track is to localize key information of pre-defined categories (field types) in the document. It is derived from the task of \emph{Key Information Localization and Extraction}, as defined in~\cite{skalicky2022business} and motivated in Section \ref{sec:related_datasets}. 

We focus the challenge on detecting semantically important values corresponding to tens of different field types rather than fine-tuning the underlying text recognition.
Towards this focus, we provide word-level text detections for each document, we choose an evaluation metric (below) that does not pay attention to the text recognition part, and we simplify the task in the challenge by only requiring correct localization of the values in the documents in the primary metric.
Text extractions are checked, besides the locations and field types, in a separate evaluation --- the leaderboard ranking does not depend on it. 
Any post-processing of values --- deduplication, converting dates to a standardized format etc., despite being needed in practice, is not performed.
With the simplifications, the main task can also be viewed as a detection problem. Note that when several instances of the same field type are present, all of them should be detected.

\begin{figure}[tb]
    \centering
    \includegraphics[width=0.6\textwidth]{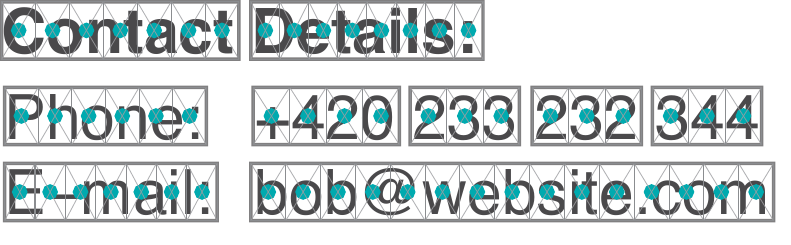}
  	\caption{Each word is split uniformly into pseudo-character boxes based on the number of characters. Pseudo-Character Centers are the centers of these boxes.}
  	\label{fig:pcc}
\end{figure}

\begin{figure}[tb]
\centering
\begin{subfigure}{.5\linewidth}
  \centering
  \includegraphics[height=40px]{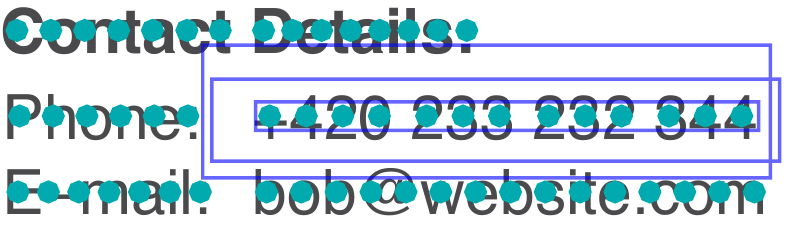}
  \caption{Correct extraction examples.}
  \label{fig:kile-metric-correct}
\end{subfigure}%
\begin{subfigure}{.5\linewidth}
  \centering
  \includegraphics[height=40px]{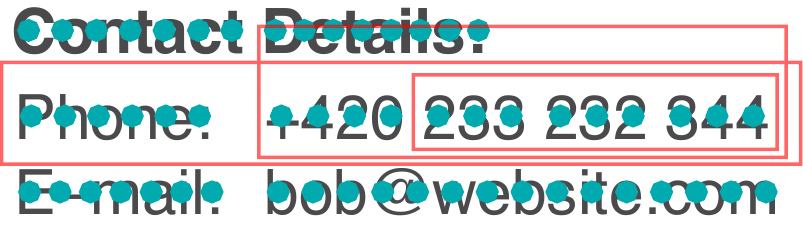}
  \caption{Incorrect extraction examples.}
  \label{fig:kile-metric-incorrect}
\end{subfigure}
\caption{Visualization of correct and incorrect bounding box predictions to capture the phone number. Bounding box must include exactly the Pseudo-Character Centers that lie within the ground truth annotation. Note: in~\ref{fig:kile-metric-correct}, only one of the predictions would be considered correct if all three boxes were predicted.}
\label{fig:kile-metrics-compare}
\end{figure}

\vspace{-8pt}
\subsubsection{The Challenge Evaluation Metric.}
Since the task is framed as a detection problem, the standard \emph{Average Precision} metric is used as the main evaluation metric.
Unlike the common practice in object detection, where true positives are determined by thresholding the Intersection-over-Union, we use a different criterion tailored to evaluate the usefulness of detections for text read-out. 
Inspired by the CLEval metric~\cite{baek2020cleval} used in text detection, we measure whether the predicted area contains nothing but the related character centers. Since character-level annotations are hard to obtain, we use CLEval definition of Pseudo-Character Center (PCC), visualized in Figure~\ref{fig:pcc}. Examples of correct and incorrect detections are depicted in Figure~\ref{fig:kile-metrics-compare}.

\subsection{Track 2: Line Item Recognition}
\label{sec:track2_lir}
The goal of the second track is to localize key information of pre-defined categories (field types) and group it into line items~\cite{denk2019bertgrid,holevcek2019table,palm2017cloudscan,majumder2020representation,BenschPS21}. A \emph{Line Item} (LI) is a tuple of fields (e.g., \emph{description}, \emph{quantity}, and \emph{price}) describing a single object instance to be extracted, e.g., a row in a table, as visualized in Figure~\ref{fig:example_document} and explained in Section \ref{sec:related_datasets}.

\vspace{-10pt}
\subsubsection{The Challenge Evaluation Metric.}
The main evaluation metric is the micro F1 score over all line item fields. A predicted line item field is correct if it fulfills the requirements from Track~1 (on field type and location) and if it is assigned to the correct line item. Since the matching of ground truth (GT) and predicted line items may not be straightforward due to errors in the prediction, our evaluation metric chooses the best matching in two steps:
\begin{enumerate}[topsep=1pt]
    \item for each pair of predicted and GT line items, the predicted fields are evaluated as in Track~1,
    \item the maximum matching is found between predicted and GT line items, maximizing the overall recall.
\end{enumerate}

\subsection{Benchmark Dataset Rules}
\label{subsec:benchmark-dataset-rules}

The use of external document datasets (and models pre-trained on such datasets) is prohibited in the benchmark in order to focus on clear comparative evaluation of methods that use the provided collection of labeled and unlabeled documents. Usage of datasets and pre-trained models from other domains, such as images from ImageNet~\cite{russakovsky2015imagenet} or texts from BooksCorpus~\cite{zhu2015aligning}, is allowed.

\section{Baseline Methods}
\label{sec:baselines}
We provide as baselines several popular state-of-the-art transformer architectures, covering text-only (RoBERTa), image-only (DETR) and multi-modal (LayoutLMv3) document representations. The code and model checkpoints for all baseline methods are distributed with the dataset.

\subsection{Multi-label NER Formulation for KILE \& LIR Tasks}
\label{subsec:multi-label-model}

The baselines described in Sections \ref{sec:roberta} and \ref{sec:layoutlmv3} use a joint multi-label NER formulation for both KILE and LIR tasks. 
The LIR task requires not only to correctly classify tokens into one of the LIR classes, but also the assignment of tokens to individual Line Items. For this purpose, we add classes $<$B-LI$>$, $<$I-LI$>$, $<$O-LI$>$ and $<$E-LI$>$, representing the beginning of the line item, inside and outside tokens and end token of the line item.
We found it is crucial to re-order the OCR tokens in top-down, left-to-right order for each predicted text line of the document. 
We provide a detailed description of the OCR tokens re-ordering in the Supplementary Material. 
For the LIR and KILE classification, we use the standard BIO tagging scheme. We use the binary cross entropy loss to train the model. 

The final KILE and LIR predictions are formed by the merging strategy as follows. We group the predicted tokens based on the membership to the predicted line item (note that we can assign the tokens which do not belong to any line item to a special group $\emptyset$), then we use the predicted OCR text lines to perform the horizontal merging of tokens assigned to the same class. Next, we construct a graph from the horizontally merged text blocks, based on the thresholded $x$ and $y$ distances of the text block pairs (as a threshold we use the height of the text block with a $25\%$ margin). The final predictions are given by merging the graph components. By merging, we mean taking the union of the individual bounding boxes of tokens/text blocks and for the text value, if the horizontal merging is applied, we join the text values with space, and with the new line character when vertical merging is applied. 

Note that this merging strategy is rather simplistic and its proper redefinition might be of interest for the participants who cannot afford training of big models as we publish also the baselines model checkpoints. 

\subsection{RoBERTa}
\label{sec:roberta}

RoBERTa~\cite{liu2019roberta} is a modification of the BERT~\cite{devlin2018bert} model which uses improved training scheme and minor tweaks of the architecture (different tokenizer). It can be used for NER task simply by adding a classification head after the RoBERTa embedding layer. Our first baseline is purely text based and uses RoBERTa\textsubscript{BASE} as the backbone of the joint multi-label NER model described in Section \ref{subsec:multi-label-model}.

\subsection{LayoutLMv3}
\label{sec:layoutlmv3}
While the RoBERTa-based baseline only operates on the text input, LayoutLMv3 \cite{huang2022layoutlmv3} is multi-modal transformer architecture that incorporates image, text, and layout information jointly. The images are encoded by splitting into non-overlap\-ping patches and feeding the patches to a linear projection layer, after which they are combined with positional embeddings. The text tokens are combined with one-dimensional and two-dimensional positional embeddings, where the former accounts for the position in the sequence of tokens, and the latter specifies the spatial location of the token in the document. The two-dimensional positional embedding incorporates the layout information. All these tokens are then fed to the transformer model. We use the LayoutLMv3\textsubscript{BASE} architecture as our second baseline, also using the multi-label NER formulation from Section \ref{subsec:multi-label-model}. Since LayoutLMv3\textsubscript{BASE} was pre-trained on an external document dataset, prohibited in the benchmark, we pre-train a checkpoint from scratch in Section~\ref{sec:pretraining}.

\subsection{Pre-training for RoBERTa and LayoutLMv3}
\label{sec:pretraining}

We use the standard masked language modeling~\cite{devlin2018bert} as the unsupervised pre-training objective to pre-train RoBERTa\textsubscript{OURS} and LayoutLMv3\textsubscript{OURS}\footnote{Note that LayoutLMv3\textsubscript{BASE}~\cite{huang2022layoutlmv3} used two additional pre-training objectives, namely masked image modelling and word-patch alignment. Since pre-training code is not publicly available and some of the implementation details are missing, LayoutLMv3\textsubscript{OURS} used only masked language modelling.
} models. The pre-training is performed from scratch using the $932$k unlabeled samples introduced in Section~\ref{sec:docileDataset}. Note that the pre-training uses the OCR predictions provided with the dataset (with reading order re-ordering).

Additionally, RoBERTa\textsubscript{BASE/OURS+SYNTH} and
LayoutLMv3\textsubscript{OURS+SYNTH} baselines use supervised pre-training on the DocILE synthetic data.

\subsection{Line Item Detection via DETR}
As an alternative approach to detecting Line Items, we use the DETR~\cite{carion2020detr} object detector, as proposed for table structure recognition on the \mbox{PubTables-1M} dataset~\cite{smock2022pubtables}.  Since pretraining on other document datasets is prohibited in the DocILE benchmark, we initialize DETR from a checkpoint\footnote{\url{https://huggingface.co/facebook/detr-resnet-50}} pretrained on COCO \cite{lin2014coco}, not from~\cite{smock2022pubtables}.

Two types of detectors are fine-tuned independently.
$\text{DETR}^\text{table}$ for table detection and $\text{DETR}^\text{LI}$ for line item detection given a table crop --- which in our preliminary experiments lead to better results than one-stage detection of line items from the full page.

\begin{table}[tb]
    \centering
    \caption{Baseline results for KILE \& LIR. $\text{LayoutLMv3}\textsubscript{BASE}$, achieving the best results, was pre-trained on another document dataset -- IIT-CDIP~\cite{lewis2006building}, which is prohibited in the official benchmark. The best results among permitted models are \underline{underlined}. The primary metric for each task is shown in \textbf{bold}.}
    \resizebox{\textwidth}{!}{
    \begin{tabular}{l|cccc|cccc}
\toprule
 & \multicolumn{4}{c|}{\textbf{KILE}} & \multicolumn{4}{c}{\textbf{LIR}} \\
                            Model &             F1 &             \textbf{AP} &          Prec. &         Recall & \textbf{F1} &             AP &          Prec. &         Recall \\
\midrule
   $\text{RoBERTa}\textsubscript{BASE}$ & \underline{0.664} & 0.534 &       0.658 &    \underline{0.671} 
   & 0.686 & 0.576 &       0.695 &    0.678 \\
   $\text{RoBERTa}\textsubscript{OURS}$ & 0.645 & 0.515 &       0.634 &    0.656 
   & 0.686 & 0.570 &       0.693 &    0.678 \\
\textcolor{red}{$\text{LayoutLMv3}\textsubscript{BASE}$} {\color{red} \tiny (prohibited)} \hspace{1.7mm} & \textcolor{red}{0.698} & \textcolor{red}{0.553} &       \textcolor{red}{0.701} &    \textcolor{red}{0.694}
& \textcolor{red}{0.721} & \textcolor{red}{0.586} &       \textcolor{red}{0.746} &    \textcolor{red}{0.699} \\
$\text{LayoutLMv3}\textsubscript{OURS}$ & 0.639 & 0.507 &       0.636 &    0.641
& 0.661 & 0.531 &       0.682 &    0.641 \\
   $\text{RoBERTa}\textsubscript{BASE+SYNTH}$ & \underline{0.664} & \underline{0.539} &       0.659 &    0.669 
& \underline{0.698} & \underline{0.583} &       \underline{0.710} &    \underline{0.687} \\
   $\text{RoBERTa}\textsubscript{OURS+SYNTH}$ & 0.652 & 0.527 &       0.648 &    0.656
& 0.675 & 0.559 &       0.696 &    0.655 \\
$\text{LayoutLMv3}\textsubscript{OURS+SYNTH}$ & 0.655 & 0.512 &       \underline{0.662} &    0.648 
& 0.691 & 0.582 &       0.709 &    0.673 \\
  $\text{NER upper bound}$ &          0.946 &          0.897 &          1.000 &          0.897
&          0.961 &          0.926 &          1.000 &          0.926 \\
$\text{DETR}^\text{table} + \text{RoBERTa}\textsubscript{BASE}$ & - & - & - & -
& 0.682 & 0.560 &       0.706 &    0.660 \\
$\text{DETR}^\text{table} + \text{DETR}^\text{LI} + \text{RoBERTa}\textsubscript{BASE}$ & - & - & - & -
& 0.594 & 0.407 &       0.632 &    0.560 \\
\bottomrule
\end{tabular}
    }
    \label{table:results_kile_lir}
\end{table}

\subsection{Upper Bound for NER-based Solutions}
All our baselines use NER models with the provided OCR on input. This comes with limitations as a field does not have to correspond to a set of word tokens --- a field can contain just a part of some word and some words covering the field might be missing in the text detections. A~theoretical upper bound for NER-based methods that classify the provided OCR words is included in Table~\ref{table:results_kile_lir}. The upper bound constructs a prediction for each ground truth field by finding all words whose PCCs are covered by the field and replacing its bounding box with a union of bounding boxes of these words. Predicted fields that do not match their originating ground truth fields are discarded.

\subsection{Results}
\label{sec:results}

The baselines described above were evaluated on the DocILE test set, the results are in Table~\ref{table:results_kile_lir}. 
Interestingly, from our pre-trained models (marked $\text{\textsubscript{OURS}}$), the RoBERTa baseline outperforms the LayoutLMv3 baseline utilizing the same RoBERTa model in its backbone.  We attribute this mainly to differences in the LayoutLMv3 pre-training: 1) our pre-training used only the masked language modelling loss, as explained in Section~\ref{sec:pretraining}, 2) we did not perform a full hyper-parameter search, and 3) our pre-training performs image augmentations not used in the original LayoutLMv3 pre-training, these are described in the Supplementary Material.

Models pre-trained on the synthetic training data are marked with $\text{\textsubscript{SYNTH}}$. Synthetic pre-training improved the results for both KILE and LIR in all cases except for LIR with $\text{RoBERTa}\textsubscript{OURS+SYNTH}$, validating the usefulness of the synthetic subset.

The best results among the models permitted in the benchmark -- i.e. not utilizing additional document datasets -- were achieved by $\text{RoBERTa}\textsubscript{BASE+SYNTH}$.

\section{Conclusions}
\label{sec:conclusions}
The DocILE benchmark includes the largest research dataset of business documents labeled with fine-grained targets for the tasks of Key Information Localization and Extraction and Line Item Recognition. The motivation is to provide a practical benchmark for evaluation of information extraction methods in a domain where future advancements can considerably save time that people and businesses spend on document processing. 
The baselines described and evaluated in Section~\ref{sec:baselines}, based on state-of-the-art transformer architectures, demonstrate that the benchmark presents very challenging tasks. The code and model checkpoints for the baselines are provided to the research community allowing quick start for the future work.

The benchmark is used for a research competition hosted at ICDAR 2023 and CLEF 2023 and will stay open for post-competition submission for long-term evaluation. We are looking forward to contributions from different machine learning communities to compare solutions inspired by document layout modelling, language modelling and question answering, computer vision, information retrieval, and other approaches.

Areas for future contributions to the benchmark include different training objective statements --- such as different variants of NER, object detection, or sequence-to-sequence modelling~\cite{tang2022unifying}, or graph reasoning~\cite{sun2021spatial}; different model architectures, unsupervised pre-training~\cite{huang2022layoutlmv3,tang2022unifying}, utilization of table structure --- e.g., explicitly modelling regularity in table columns to improve in LIR; addressing dataset shifts~\cite{pampari2020unsupervised,sipka2022hitchhiker}; or zero-shot learning~\cite{kil2021revisiting}.

\vspace{-0.7em}
\small
\subsubsection*{\small Acknowledgements}
We acknowledge the funding and
support from Rossum and the intensive work of its annotation team, particularly Petra Hrdličková and Kateřina Večerková.
YP and JM were supported by Research Center for Informatics (project CZ.02.1.01/0.0/0.0/16\_019/0000765 funded by OP VVV), by the Grant Agency of the Czech Technical University in Prague, grant No. SGS20/171/OHK3 /3T/13, by Project StratDL in the realm of COMET K1 center Software Competence Center Hagenberg, and Amazon Research Award. DK was supported by grant PID2020-116298GB-I00 funded by MCIN/AE/NextGenerationEU and ELSA (GA 101070617) funded by EU.

\clearpage
\bibliographystyle{splncs04}
\bibliography{bibliography}



\section*{Supplementary material}

# Supplementary Material
# DocILE Benchmark for Document Information Localization and Extraction

## 1 Dataset Details

### 1.1 Document Preprocessing

The following pre-processing was applied to documents from the annotated set:

- (UCSF only) PDFs from UCSF contain a text *"Source: [URL in UCSF]"* on the bottom of the page. As this text would affect the competition tasks, it was removed from the PDFs. Instead, the original document ID is recorded in the dataset metadata.
- Skewed pages were detected in PDFs and images rendered from the PDF pages were deskewed automatically[1]. New PDFs were generated from the deskewed images[2], each rendered to have the longer dimension equal to 842 pixels (this corresponds to the longer side of A4 at 72 DPI).

### 1.2 Pre-computed OCR

To facilitate a quick start with the dataset, we provide pre-computed OCR. We used the DocTR [7] library with the DBNet[3] detector and the CRNN [9] recognition model, which showed good results in Document OCR in [8]. The predictions are for word-level tokens and include the geometry (bounding box), value (recognized text) and confidence. The words are grouped into blocks. For each word we also provide the snapped geometry, computed by binarizing the word image crop and removing rows/columns from the sides that contain mostly the background color. The implementation is available in the repository. The use of the provided OCR predictions by benchmark submissions is optional. The Pseudo-Character Centers from the snapped bounding boxes are used in the benchmark evaluation.

### 1.3 Document Layout Clustering

Documents in the DocILE dataset are assigned to different layout clusters, see Figure 1 for an example of a layout cluster. To find the layout clustering for the

---

[1] Using Leptonica: https://tpgit.github.io/Leptonica/skew_8c_source.html
[2] Using pdf2image: https://pypi.org/project/pdf2image/

2

annotated and unlabeled sets[3], layout clusters were first detected with the algorithm described in this section and then manually corrected[4] for the annotated set.

The clustering algorithm takes KILE and LIR fields on input. These fields were predicted by a proprietary model. Then, the following two distance functions were used to detect which clusters should be merged together.

1. Following our definition of layouts, each cluster is represented as a set of fields that can occur in the layout along its possible $x$-axis positions[5]. The distance between two clusters is then based on the number of field types in common (weighted by the abundance of the field type in the cluster) and the distance between the positions of the same field type. Positions are first normalized to align the left-most and right-most positions (accounting for different translation and scale which is especially important for scanned documents).
2. The second distance function uses values of a selected set of field types that are usually equal among documents of the same layout. Specifically information about the sender (name, address, registration ID and tax ID), payment terms and text in table headers. A cluster is then a collection of all possible values for these field types. Distance between two clusters is based on the edit distance between values of the same field type.

By combining the above distance functions, layout clusters were found in the following steps:

1. Starting with single document clusters for all documents, clusters were iteratively merged if their distance was under a selected threshold. At most 50 documents per cluster were sent for annotation.
2. After annotation, predicted fields were replaced with ground truth fields and the clustering was re-run. The result was then manually corrected.
3. To cluster the unlabeled set, field predictions were used and each document was assigned independently to the closest cluster in the annotated set when the distance was lower than a threshold. When the distance exceeded the threshold, it was assigned cluster id $-1$, representing that a corresponding cluster was not found – which happened for 18.8% of the unlabeled documents. On a random sample of 100 unlabeled documents, 72 documents were assigned to some cluster with a precision of 86%. Determining whether the 28 documents were correctly unassigned was not checked as it would require much more effort.

---

[3] For the synthetic set the clustering is implicit from the selection of the template documents

[4] Notice that full manual check was infeasible, as that would involve checking over half of million of cluster pairs for a potential merge. Instead, potential cluster merges and splits were generated using the two distance metrics described in Section 1.3.

[5] The $y$-axis is ignored as it often varies even for documents of the same layout, e.g., based on the length of the table.

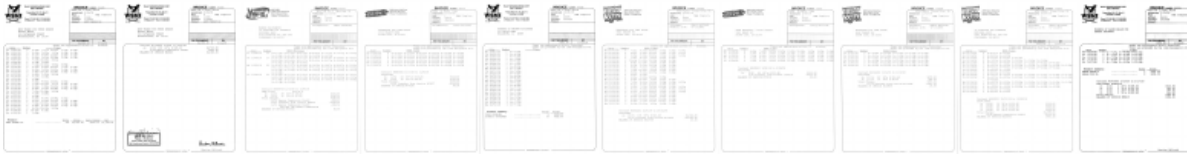

Fig. 1: First page of 10 documents belonging to the same cluster.

### 1.4 Dataset Splitting

The following rules and parameters were used to split the dataset into training, validation and test split and to select the templates for synthetic documents.

- The target number of documents in the test set and validation set is $1,000$ and $500$, respectively. Test set is selected first and then validation set is selected from the remaining documents with the same algorithm. The remaining documents form the training set. With respect to the test set, we consider the documents from both training and validation sets as *available for training*. Below we describe the selection of the test set.
- The target number of *synthetic templates* (documents from the training set whose annotations are used for generation of synthetic documents) is set to 100.
- The target ratio of documents from *PIF* and *UCSF* sources in the selected set is 55%:45%, which is similar to the ratio in the full annotated dataset. Without this constraint the distribution would be significantly different as the two sources have different distribution of cluster sizes, which affects further selection of the documents.
- Let us call a cluster $X$-shot with respect to the test set if it has $X$ documents available for training. The cluster is called zero-shot if $X = 0$, few-shot if $0 < X < 4$ and many-shot if $X \geq 4$. The target proportion of test documents belonging to $0, 1, 2, 3$-shot clusters is set to approximately $1/4, 1/12, 1/12, 1/12$, respectively, for each source. I.e., $1/2$ of the test documents belong to zero-shot or few-shot clusters and $1/2$ of the documents belong to many-shot clusters.
- To ensure diversity of test samples belonging to zero-shot and few-shot clusters, we allow at most 20 samples from each zero-shot or few-shot cluster in the test set.
- When selecting few-shot samples, approximately half of the documents should belong to clusters containing a synthetic template, e.g., $1/24 \cdot 1000 \approx 42$ of the documents in the test set should belong to clusters that contain a synthetic template and that have two documents available for training. The remaining synthetic templates are from many-shot clusters or clusters not contained in the test set.

### 1.5 Description of Annotations

Field Types for KILE and LIR are described in Tables 1 and 2, respectively. Note that some rather rare fields are not represented in the validation set: iban, bic, customer_tax_id and line_item_tax_rate. Their counts in the training set are 5, 31, 40 and 96 respectively. With the micro-averaged evaluation metric their impact on the evaluation is only minor.

The number of documents and clusters containing individual fieldtypes are displayed in Figures 2 and 3 resperctively.

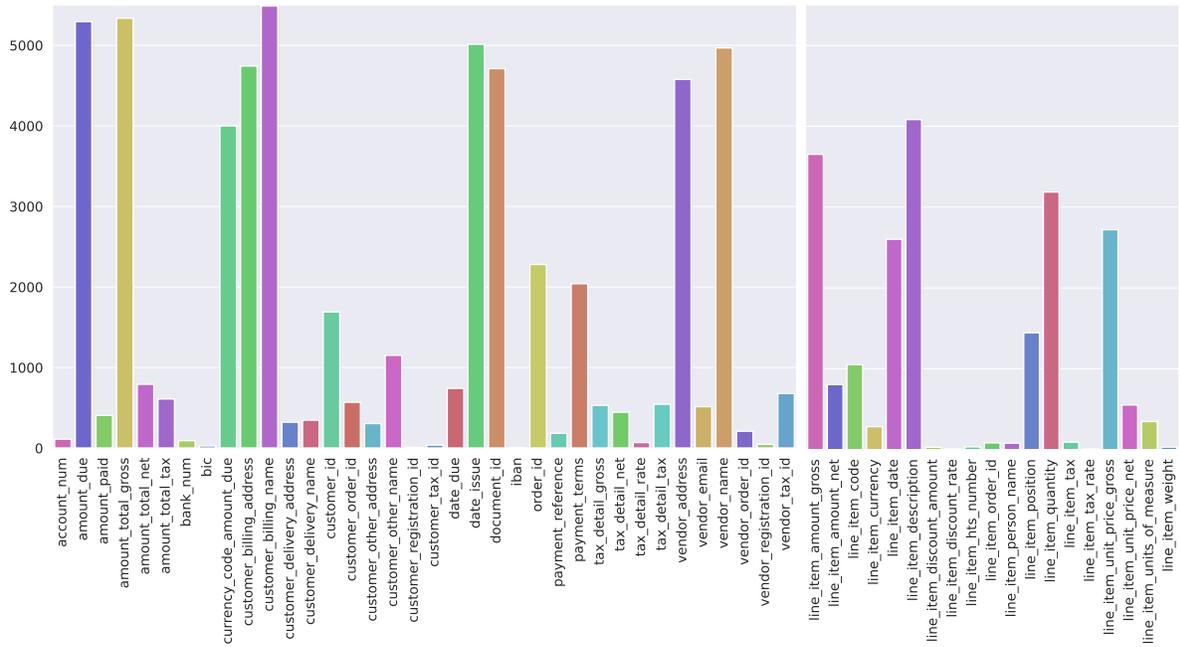

Fig. 2: The number of documents containing individual KILE (left) and LIR (right) fieldtypes, out of 5580 documents in the training and validation sets.

Table 1: Description of all KILE field types.

| field type | description |
|---|---|
| account_num | Bank account number |
| amount_due | Total amount to be payed |
| amount_paid | Total amount already paid |
| amount_total_gross | Total amount with tax |
| amount_total_net | Total amount without tax |
| amount_total_tax | Total sum of tax amounts |
| bank_num | Bank number |
| bic | Bank Identifier Code (SWIFT) |
| currency_code_amount_due | Currency code or symbol found near the amount due |
| customer_billing_address | Address of the company that is being invoiced |
| customer_billing_name | Name of the company that is being invoiced |
| customer_delivery_address | Address of the company for delivery of goods/services |
| customer_delivery_name | Name of the company for delivery of goods/services |
| customer_id | Customer account number |
| customer_order_id | Any customer order reference |
| customer_other_address | Any other name and address of the purchasing company |
| customer_other_name | Any other name of the purchasing company |
| customer_registration_id | Purchaser registration identifier number |
| customer_tax_id | Purchaser tax identification number |
| date_due | Due date for payment |
| date_issue | Date the document was issued |
| document_id | Main document number |
| iban | International Bank Account Number |
| order_id | Any order number |
| payment_reference | Payment reference number |
| payment_terms | Conditions for the payment time window |
| tax_detail_gross | Tak breakdown line amount with tax |
| tax_detail_net | Tax breakdown line amount without tax |
| tax_detail_rate | Tax breakdown line tax rate |
| tax_detail_tax | Tax breakdown line tax amount |
| vendor_address | Address of the supplier company |
| vendor_email | Any supplier e-mail address |
| vendor_name | Name of the supplier company |
| vendor_order_id | Any vendor order reference |
| vendor_registration_id | Supplier registration identification number |
| vendor_tax_id | Supplier tax identification number |

Table 2: Description of all LIR field types.

| field type | description |
| --- | --- |
| line_item_amount_gross | Total amount with tax for item |
| line_item_amount_net | Total amount without tax for item |
| line_item_code | Item article number |
| line_item_currency | Line currency (if standalone) |
| line_item_date | Date (e.g. item delivery date) |
| line_item_description | Goods and services description |
| line_item_discount_amount | Total discount amount |
| line_item_discount_rate | Discount rate |
| line_item_hts_number | Harmonized Tariff Schedule number |
| line_item_order_id | Related order reference number |
| line_item_person_name | Person name (e.g. who performed the service) |
| line_item_position | Line index, order of items |
| line_item_quantity | Quantity |
| line_item_tax | Line tax amount |
| line_item_tax_rate | Tax rate (percentage, verbal) |
| line_item_unit_price_gross | Price with tax per unit |
| line_item_unit_price_net | Price without tax per unit |
| line_item_units_of_measure | Unit of measure |
| line_item_weight | Item(s) weight (net, gross) |

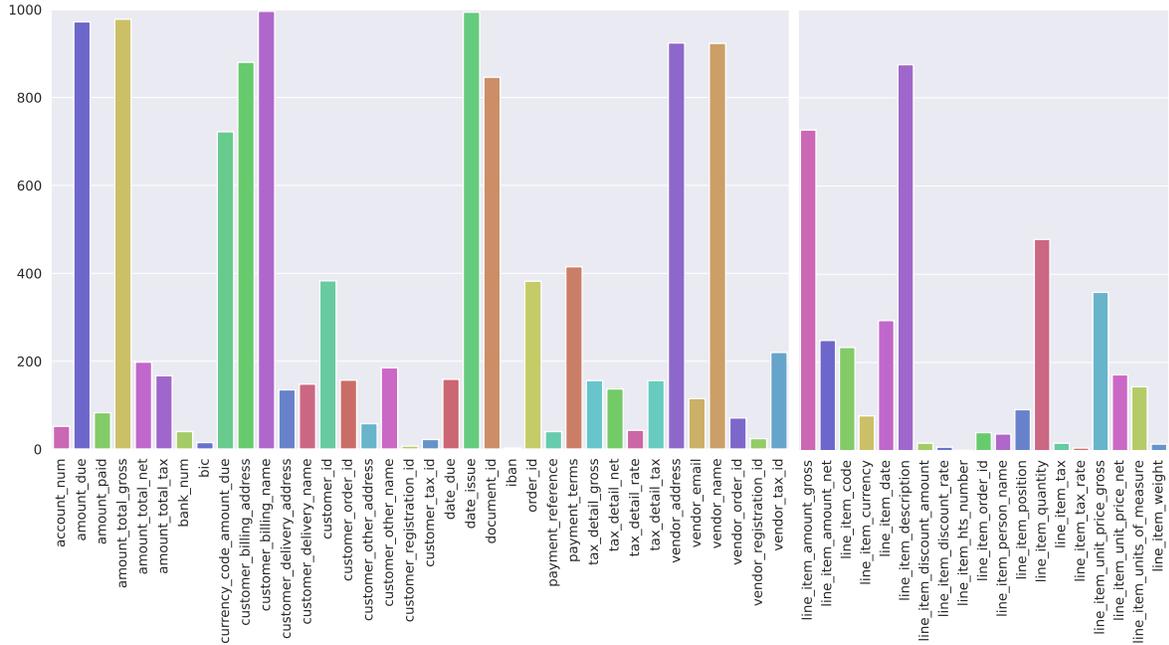

Fig. 3: The number of clusters containing individual KILE (left) and LIR (right) fieldtypes, out of 1063 clusters in the training and validation sets.

Table 3: Evaluation on zero/few/many-shot subsets of the test set for the RoBERTa$_{BASE}$ baseline for KILE and LIR based on the number of documents of the same layout cluster available for training (in train.+val. set).

| Task | Model | Training size | F1 | AP | Prec. | Recall |
|---|---|---|---|---|---|---|
| KILE | RoBERTa$_{BASE}$ | 0 | 0.524 | 0.394 | 0.504 | 0.547 |
| KILE | RoBERTa$_{BASE}$ | 1-3 | 0.620 | 0.483 | 0.621 | 0.619 |
| KILE | RoBERTa$_{BASE}$ | 4+ | 0.742 | 0.609 | 0.742 | 0.742 |
| LIR | RoBERTa$_{BASE}$ | 0 | 0.598 | 0.504 | 0.568 | 0.630 |
| LIR | RoBERTa$_{BASE}$ | 1-3 | 0.568 | 0.406 | 0.575 | 0.560 |
| LIR | RoBERTa$_{BASE}$ | 4+ | 0.756 | 0.643 | 0.789 | 0.726 |

## 2  Evaluation Details

### 2.1  Additional Evaluation Metrics

Although the two benchmark tasks use different primary metrics, the evaluation of both tasks includes AP, F1, precision and recall. Predictions can use an optional flag `use_only_for_ap` to indicate the prediction does not have big enough confidence and should not be counted towards F1, precision and recall but it should still be used for AP[6]. When AP is computed for LIR, only the "confident" predictions (with `use_only_for_ap=False`) are used to find the perfect matching between line items.

Additionally, we set up a secondary evaluation benchmark for end-to-end KILE and LIR, where a correctly recognized field also needs to exactly read out the text.

The benchmark also computes results on the test subsets to compare the performance on the 0-shot, few-shot and many-shot samples (based on how many documents from the same layout cluster are available for training) as shown in Table 3. Furthermore, it is possible to evaluate separately on samples belonging to clusters for which synthetic documents were generated, to study the influence of using synthetic data. The provided evaluation code supports evaluation on any dataset subset, e.g., based on the source of the document (*UCSF* vs *PIF*).

### 2.2  AP Implementation Details

In Average Precision, predictions are sorted by score (confidence) from the highest to the lowest and added iteratively. This ordering is not well defined if several predictions have the same score or if the scores are not provided. Taking into account the `use_only_for_ap` flag, effectively marking some predictions as less confident, the predictions are sorted by the following criteria:

---

[6] Note that when computing AP, adding predictions with lower score than all other predictions can never decrease the metric, no matter how low is the precision among these extra predictions.

1. Predictions with `use_only_for_ap=False` go first.
2. Predictions with higher score go first.
3. Predictions are added in the provided order (*prediction document rank*), i.e., start with the first prediction for each document, then take the second prediction for each document, etc.
4. To break the tie for predictions with the same score and prediction document rank, the document id is hashed along the prediction document rank. This ensures a deterministic but different ordering of documents for each prediction document rank, preventing some documents to have higher influence on the final result.

Other implementation details for AP have been done in the same way as in the standard COCO[4] evaluation. Namely:

- When the Precision-Recall curve has a zig-zag pattern (precision increases for higher recall), the gaps are filled to the left.
- For two consecutive (recall, precision) pairs $(r_1, p_1)$, $(r_2, p_2)$ where $r_2 > r_1$ we use the precision $p_2$ for the interval $[r_1, r_2]$ when computing the Average Precision.

This can be also explained as computing the area under a function (for recall between 0 to 1) $precision(r)$ defined as:

$$precision(r) \coloneqq \max\{p' | \text{there exists } (p', r') \text{ with } r' \geq r\}$$

### 2.3 Updated Results

We note that the results in the main paper and in the supplementary material were updated after the first version of this arXiv pre-print, to reflect updates in the published code (Section 4).

## 3 Baseline Details

**Unsupervised Pre-training:** The RoBERTa$_{\text{OURS}}$[5] model was pre-trained for 50k training steps with a batch size of 64. The LayoutLMv3$_{\text{OURS}}$[1] was pre-trained using AdamW optimizer [6] for 30 epochs using a batch size of 16. Both trainings use a cosine decay for the learning rate with a linear warmup. Note that LayoutLMv3 [1] uses three training objectives: masked language modeling, masked image modeling, and word-patch alignment loss. Whereas our pre-training only uses masked language modeling. Furthermore, LayoutLMv3 pre-trains without any data augmentations on the IIT-CDIP [2] dataset. Our setup uses random horizontal flipping of the images and trains on the unlabeled subset of the introduced DocILE dataset.

**Supervised Pre-training on Synthetic Documents:** We run 30 epochs of supervised pre-training on DocILE synthetic dataset for RoBERTa$_{\text{BASE}}$, RoBERTa$_{\text{OURS}}$, and LayoutLMv3$_{\text{OURS}}$ backbones of the joint multi-label NER model, with the following parameters: learning rate 2e−5, weight decay 0.001, batch size 16.

**Algorithm 1** OCR re-ordering algorithm

---

**Require: b**                                                                              ▷ OCR word bboxes
  **function** GET_CENTER_LINE_CLUSTERS(**b**)
    *compute histograms of heights and centroids*
    *create clusters based on the centroids and height histograms*
  **end function**

  **function** SPLIT_FIELDS_BY_TEXT_LINES(**b**)
    **c** = **get_center_line_clusters**(**b**)
    **for** $b \in \mathbf{b}$ **do**
      $b_{\text{line\_id}} \leftarrow \arg\min_{c \in \mathbf{c}} |b_y - c_y|$                        ▷ Assign line numbers
    **end for**
  **end function**

  **function** GET_SORTED_FIELD_CANDIDATES(**b**)
    **b** ← **split_fields_by_text_lines**(**b**)
    **sort**(**b**)                                                           ▷ Sort by assigned line numbers
    **l** ← **group**($\mathbf{b}_{\text{line\_id}}$)
    $\forall l \in \mathbf{l} : \mathbf{sort}(l_x)$                                     ▷ Sort by $x$-coordinate
  **end function**

---

**Supervised Training on Annotated Documents:** All RoBERTa and LayoutLMv3 were trained for 750 epochs with learning rate 2e−5, weight decay 0.001 and batch size[7] 16.

For DETR, we train the model using the Adam optimizer with learning rate $3 \cdot 10^{-5}$ for the transformer and $3 \cdot 10^{-7}$ for the convolutional backbone (ResNet-50), weight decay 1e-4, FP16 precision, batch size 32 and early stopping on the validation loss.

**OCR Re-ordering** We observed that for the Line Item separation task, it is crucial to provide the text tokens in per-line reading order (i.e., from top to bottom, each text line from left to right). The pseudocode of the used re-ordering algorithm is in Algorithm 1. Since we use a joint multi-label model for both KILE and LIR, the re-ordering was applied for every training and inference.

## 4  Code and Dataset Download

For the dataset download instructions, baseline implementations and checkpoints, we refer the reader to the https://github.com/rossumai/docile repository.

---

[7] Since we used multiple GPUs that had different memory sizes, for some trainings we had to decrease the batch size. In these cases we appropriately increased the gradient accumulation step.

10



\end{document}